\documentclass[lettersize,journal]{IEEEtran}
\usepackage{amsmath,amsfonts}
\usepackage{algorithmic}
\usepackage{algorithm}
\usepackage{array}
\usepackage[caption=false,font=normalsize,labelfont=sf,textfont=sf]{subfig}
\usepackage{xcolor}
\usepackage{textcomp}
\usepackage{stfloats}
\usepackage{url}
\usepackage{verbatim}
\usepackage{graphicx}
\usepackage{cite}
\usepackage{tabularx}
\usepackage{booktabs}
\usepackage{graphicx}
\usepackage{array}
\usepackage[numbers]{natbib}
\usepackage{caption}
\usepackage{multirow}
\usepackage{booktabs}
\usepackage{array}
\usepackage{xcolor}

\usepackage[font=footnotesize]{caption}
\begin{document}
\title{A Multi-view Divergence-Convergence Feature Augmentation Framework for Drug-related Microbes Prediction}

\author{Xin An$^{1, 6\dagger}$, \text{Ruijie Li}$^{1,3,6\dagger}$~\IEEEmembership{Student Member, IEEE}, \text{Qiao Ning}$^{1,2,5*}$, \text{Shikai Guo}$^{1,3,6}$, \text{Hui Li}$^{1}$, \text{Qian Ma}$^{1}$\thanks{

1 Department of Information Science and Technology, Dalian Maritime University, Dalian, 116026, P.R. China.

2 School of Artificial Intelligence and Computer Science, Jiangnan University, Wuxi, 214122, P.R. China.

3 DUT Artificial Intelligence Institute, Dalian, P.R. China.

4 School of Computer Science and Engineering, Northeastern University, Shenyang, 110819, P.R. China.

5 Key Laboratory of Symbolic Computation and Knowledge Engineering of Ministry of Education, Jilin University, Changchun, 130012, P.R. China

6 Dalian Key Laboratory of Artificial Intelligence, Dalian, P.R. China.

* To whom correspondence should be addressed at ningq669@jiangnan.edu.cn

\text{†} Xin An and Ruijie Li contributed equally to this work and should be considered co-first authors.}}

% The paper headers
\markboth{Journal of \LaTeX\ Class Files,~Vol.~14, No.~8, August~2024}%
{Shell \MakeLowercase{\textit{et al.}}: A Sample Article Using IEEEtran.cls for IEEE Journals}

\IEEEpubid{0000--0000/00\$00.00~\copyright~2024 IEEE}
% Remember, if you use this you must call \IEEEpubidadjcol in the second
% column for its text to clear the IEEEpubid mark.

\maketitle

\begin{abstract}
In the study of drug function and precision medicine, identifying new drug-microbe associations is crucial. 
% Current methods insufficiently explore drug-microbe associations and fail to deeply integrate and learn their similarity relationships. 
However, current methods isolate association and similarity analysis of drug and microbe, lacking effective inter-view optimization and coordinated multi-view feature fusion.
In our study, a multi-view \textbf{D}ivergence-\textbf{C}onvergence \textbf{F}eature \textbf{A}ugmentation framework for \textbf{D}rug-related \textbf{M}icrobes \textbf{P}rediction (DCFA\_DMP) is proposed, to better learn and integrate association information and similarity information. In the divergence phase, DCFA\_DMP strengthens the complementarity and diversity between heterogeneous information and similarity information by performing Adversarial Learning method between the association network view and different similarity views, optimizing the feature space. In the convergence phase, a novel Bidirectional Synergistic Attention Mechanism is proposed to deeply synergize the complementary features between different views, achieving a deep fusion of the feature space. Moreover, Transformer graph learning is alternately applied on the drug-microbe heterogeneous graph , enabling each drug or microbe node to focus on the most relevant nodes. Numerous experiments demonstrate DCFA\_DMP’s significant performance in predicting drug-microbe associations. It also proves effectiveness in predicting associations for new drugs and microbes in cold start experiments, further confirming its stability and reliability in predicting potential drug-microbe associations.
\end{abstract}

\begin{IEEEkeywords}
Drug-related Microbes Prediction, Divergence-Convergence, Adversarial Learning, Bidirectional Synergistic Attention Mechanism
\end{IEEEkeywords}

\section{Introduction}
\label{tab:Introduction}
\IEEEPARstart{A}{t} birth, humans consist exclusively of their own eukaryotic cells. However, during the early years of life, skin, mouth, and gut are colonized by a diverse array of bacteria, archaea, fungi, and viruses. This community of cells is referred to as the human microbiome~\citep{morgan2012chapter}. Dysbiosis within the microbiome has been linked to various diseases, including inflammatory bowel disease, multiple sclerosis, diabetes (type 1 and type 2), allergies, asthma, autism, and cancer~\citep{lloyd2016healthy}. Early in drug discovery, it was recognized that the microbiota plays a crucial role in the efficacy of therapeutic compounds. However, with the ongoing emergence of antimicrobial resistance, there is an urgent need to systematically determine the interactions between microbes and drugs to facilitate drug development.

Therefore, discovering novel microbiome-drug associations is of great significance in drug efficacy studies and precision medicine. An increasing body of literature has reported the links between microbes and drugs. For example, Zimmermann et al.~\citep{zimmermann2019mapping} found that the activity and toxicity of many oral drugs can be altered by bacterial enzymes produced by the human gut microbiota, potentially influencing treatment outcomes. Among these, the gut bacterium \textit{Bacteroides thetaiotaomicron} is a prolific drug metabolizer capable of metabolizing a wide range of drugs, including diltiazem. Furthermore, Blaser et al.~\citep{blaser2009consequences} suggested that the skin microbiome plays a crucial role in antibiotic resistance in \textit{Staphylococcus aureus} infections. These findings underscore the importance of systematically elucidating microbe–drug interactions to advance drug development and address the challenges posed by antimicrobial resistance.

\IEEEpubidadjcol
In recent years, graph learning has become increasingly prevalent in predicting microbiome-drug associations due to its remarkable capacity to model graph-structured data. Several variants, including Graph Convolutional Networks (GCN~\citep{kipf2016semi})-based approaches~\citep{long2020predicting, huang2023gnaemda, wang2023microbe}, Graph Attention Networks (GAT~\citep{velickovic2017graph})-based methods~\citep{long2020ensembling, liu2021mgatmda}, and Heterogeneous Graph Attention Networks (HAN~\citep{wang2019heterogeneous})-based models~\citep{xing2024mdmd} showing considerable promise in this task.

For instance, NGMDA~\citep{xuan2024multi} builds a drug-microbe heterogeneous graph that encodes the positional and topological features of drug and microbe nodes, combining different feature types from neighboring nodes and the entire heterogeneous graph. DHDMP~\citep{xuan2024dynamic} constructs a hypergraph with dynamic topology, utilizing GCN to transfer node features to the heterogeneous graph, thereby enhancing neighbor feature representation learning. MKGCN~\citep{cui2023mkgcn} extracts features from microbes and drugs through multiple GCN layers, using kernel matrices to predict their associations. GACNNMDA~\citep{ma2023gacnnmda} integrating GAT and CNN to predict microbiome-drug interactions using heterogeneous networks. Although these methods have shown promising results, they still have limitations. NGMDA overlooks the complex relationships among multiple drugs and microbes. DHDMP cannot fully capture certain intricate long-distance correlations. MKGCN fails to capture complex structural and semantic relationships in heterogeneous networks. Overall, although these graph learning-based models have shown promising results in drug-microbe prediction tasks, they still cannot fully leverage the information within heterogeneous networks and similarity networks.

With the development of deep learning, an increasing number of researchers have proposed a series of computational methods based on deep learning to detect potential microbiome-drug interactions~\citep{zhao2022predicting, algavi2023data}, achieving promising results. For instance, Zhu et al.~\citep{zhu2019prediction} introduced a method called HMDAKATZ, which employs the KATZ metric to infer potential microbiome-drug associations. 
Long et al.~\citep{long2020association} proposed the HNERMDA framework, which integrates metapath2vec with bipartite network recommendations to predict potential microbiome-drug interactions.
Additionally, Zhu et al.~\citep{zhu2021predicting} designed a computational model called LRLSMDA, which uses the Laplacian regularized least squares algorithm to identify microbiome-drug associations. 
Tan et al.~\citep{tan2022gsamda} proposed the GSAMDA model, which employs GAT-based and sparse autoencoder modules to learn both topological and attribute representations of nodes in heterogeneous networks. 
% However, these computational methods often rely on simple metrics to evaluate the strength of associations between microbes and drugs, and there is still room for improvement in learning the features of heterogeneous networks.
However, existing computational approaches typically suffer from isolated processing of association networks and similarity graphs, lacking optimization mechanisms to enhance informational diversity. Furthermore, their feature fusion mechanisms fail to establish bidirectional synergistic relationships between divergent feature representations, ultimately resulting in suboptimal integration of multi-source information.

% Numerous studies have investigated the associations between microorganisms and drugs, leading to the development of various computational methods aimed at detecting potential microbe–drug interactions with promising outcomes. For example, Zhu et al.~\cite{zhu2019prediction}introduced a method named HMDAKATZ, which utilizes the KATZ metric to infer potential microbe–drug associations. Long et al.~\cite{long2020association} designed a computational framework called HNERMDA, which predicts possible microbe–drug interactions by integrating metapath2vec with bipartite network recommendation techniques. Additionally, Zhu et al.~\cite{zhu2021predicting} proposed a computational model named LRLSMDA, which identifies microbe–drug associations based on the Laplacian Regularized Least Squares algorithm. Despite these advancements, existing computational methods often rely on simplistic metrics to evaluate the strength of associations between microbes and drugs and exhibit limitations in effectively processing and learning features from heterogeneous networks.Consequently, there remains significant potential for improving these approaches to enhance the accuracy and comprehensiveness of microbe–drug association predictions.

\begin{figure*}
\centering
\includegraphics[width=\textwidth]{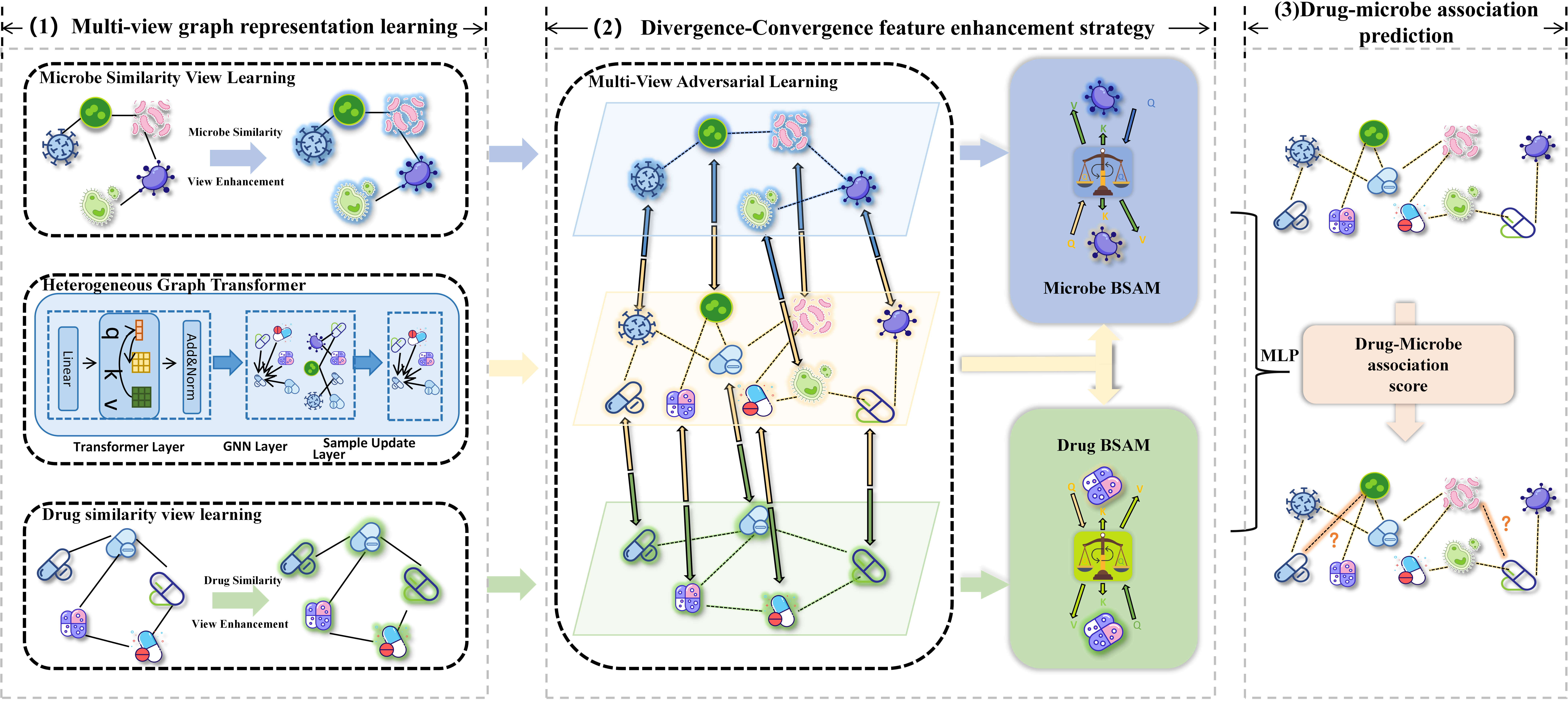}
\captionsetup{font=scriptsize}
\caption{Workflow of DCFA\_DMP: (1)  Multi-view graph representation learning; (2) Divergence-Convergence feature enhancement strategy; (3) Drug-microbe association prediction; }
\label{fig:1}
\end{figure*}

To address these challenges, we propose a multi-view divergence-convergence feature augmentation framework for drug-related microbes prediction, whose framework is shown in Figure~\ref{fig:1}. The contributions of this method are as follows:

\begin{itemize}
\item We propose a transformer-based graph learning approach, where Transformer extends the receptive field of the GNN, facilitating a more comprehensive exploration of drug-microbe associations and GNNs capture essential structural information, guiding the Transformer to focus on key local features.
\item We introduce a novel divergence-convergence feature enhancement framework. In the divergence phase, an Adversarial Learning approach is employed to enhance the complementarity and diversity of information between the association and similarity views. In the convergence phase, we propose a novel Bidirectional Synergistic Attention Mechanism (BSAM) to facilitate multi-dimensional interaction and fusion of the enhanced multi-view features.
\item Extensive experiments, including cold start and case studies, show that the proposed method outperforms baseline models on benchmark datasets and proves the practicality of our model in drug localization tasks. 
\end{itemize}

\section{Related Materials}

\section{Methodology}

The framework consists of three main components: 1) Multi-view graph representation learning;
% , which includes a feature convolution module and a Transformer-based graph learning module, aimed at learning graph representation embeddings from multiple view perspectives. The feature convolution module learns the similarity network information for drugs and microbes, while the Transformer-based graph learning module captures features from the drug-microbe heterogeneous network. The Transformer-based graph learning module sequentially includes three parts: (i) Transformer layer; (ii) GNN layer; (iii) Sample update layer. The Transformer layer is used to expand the receptive field of the GNN layer and efficiently aggregate attention-based sample information, while the GNN layer helps the Transformer layer perceive graph structural information and obtain more relevant information from neighboring nodes. The sample update layer is integrated to efficiently update attention-based samples. 
2) Divergence-Convergence feature enhancement strategy;
% , utilizing similarity and association graphs for contrastive learning and cross-attention to enhance drug and microbe feature extraction. 
3) Drug-microbe association prediction. In the subsequent sections, we will provide a comprehensive review of each stage and present the essential details pertaining to our model.

\subsection{Multi-view graph representation learning}

\subsubsection {Feature convolution module}
Different similarity relationships can reflect the diverse semantic connections between drug and microbe nodes. Therefore, we constructed a K-nearest neighbor graph (KNN) based on the similarity matrix to capture the underlying structure of drugs and microbes in the feature space, and to learn embeddings that capture specific information.

We denote \( X^d \) as the drug similarity matrix, and construct the binary adjacency matrix \( A^d \) of the drug KNN graph based on the similarity between each pair of drugs. The specific definition of each entry \( A_{ij}^d \) in the matrix \( A^d \) is as follows:
\begin{equation} 
A_{ij}^d = 
\begin{cases} 
1, & \text{if } d_j \in \tilde{N}_k(d_i) \\ 
0, & \text{otherwise} 
\end{cases} 
\end{equation}
where \( \tilde{N}_k(d_i) = \{d_i\} \cup N_k(d_i) \) represents the extended k-neighborhood set of drug \( d_i \), where \( N_k(d_i) \) denotes the k-nearest neighbors of drug \( d_i \). 

Similarly, we denote \( X^m \) as the microbe similarity matrix, and the specific definition of each entry \( A_{ij}^m \) in the matrix \( A^m \) is as follows:
\begin{equation} 
A_{ij}^m = 
\begin{cases} 
1, & \text{if } m_j \in \tilde{N}_k(m_i) \\ 
0, & \text{otherwise} 
\end{cases} 
\end{equation}
where \( \tilde{N}_k(m_i) = \{m_i\} \cup N_k(m_i) \) represents the extended k-neighborhood set of microbe \( m_i \), where \( N_k(m_i) \) denotes the k-nearest neighbors of microbe \( m_i \).

In the feature space, we use the drug KNN graph \( G_r = (A^d, X^d) \) and the microbe KNN graph \( G_d = (A^m, X^m) \), and employ the classical GCN to represent the output of the \( l \)-th layer of the constructed graphs as follows:
\begin{equation}
\begin{cases}
H_d^{(l)} = \text{ReLU}(D_d^{-\frac{1}{2}} A^d D_d^{-\frac{1}{2}} H_d^{(l-1)} W_d^{(l)}) \\
H_m^{(l)} = \text{ReLU}(D_m^{-\frac{1}{2}} A^m D_m^{-\frac{1}{2}} H_m^{(l-1)} W_m^{(l)})
\end{cases}
\end{equation}
where \( H_d^{(l)} \) and \( H_m^{(l)} \) represent the feature matrices at the \( l \)-th layer for drug and microbe propagation. ReLU denotes the ReLU activation function. \( D_d \) and \( D_m \) are diagonal matrices, with each element representing the degree of the corresponding node. \( W_d^{(l)} \) and \( W_m^{(l)} \) are the weight matrices at the \( l \)-th layer of the GCN for drugs and microbes, respectively.

\subsubsection {Transformer Graph Learning Module}
Traditional GNN often rely on local neighborhood information to learn complex relationships between drugs and microbes, limiting their ability to capture long-range relationships. Additionally, as the network depth increases, information tends to be over-smoothed during propagation, reducing in the discriminative power of node representations and failing to effectively capture the intricate features of drug-microbe heterogeneous graphs. Although Transformer capture long-range dependencies, they typically neglect certain structural information in complex graph-structured data, which increases computational complexity. To address these challenges, we combine GNN with Transformer to leverage the strengths of both.

The module consists of three parts: the transformer layer, the GNN layer, and the sample update layer. The transformer layer expands the receptive field of the GNN layer, allowing it to aggregate more comprehensive drug-microbe association information; the GNN layer helps the transformer layer perceive graph structural and obtain more relevant information from adjacent drug-microbe nodes; the sample update layer updates the attention samples for drug-microbe pairs.

\textbf{Transformer layer: }The transformer layer extends the receptive field of the GNN layer, enabling it to focus on potential, important nodes that are farther apart. To reduce computational complexity and avoid irrelevant information, we focus only on the most relevant nodes for each center node.

First, for each node \( v_i \)'s input \( H_{\text{Smp}_{(v_i)}} \), we use three matrices \( W_{\text{query}} \), \( W_{\text{key}} \), and \( W_{\text{value}} \) to map the node representations to the query \( Q \), key \( K \), and value \( V \). Based on the attention mechanism, these values are aggregated as follows:
\begin{equation} 
h_i = \text{softmax}\left(\frac{QK^T}{\sqrt{d_k}}\right)V
\label{eqn:hi}
\end{equation} 
where
% \begin{equation}
% \begin{cases}
% Q = h_i W_{\text{query}} \\
% K = H_{\text{Smp}_{(v_i)}} W_{\text{key}} \\
% V = H_{\text{Smp}_{(v_i)}} W_{\text{value}}
% \end{cases}
% \end{equation}
\( Q=H_{\text{Smp}_{(v_i)}} \cdot W_{\text{query}}\) represents the query vector, \( K=H_{\text{Smp}_{(v_i)}} \cdot W_{\text{key}}\) represents the key vector, \( V=H_{\text{Smp}_{(v_i)}} \cdot W_{\text{value}}\) represents the value vector.
% and the matrix \( H_{\text{Smp}_{(v_i)}} \) represents the input information of node \( v_i \). 
To learn richer contextual information from multiple subspaces, the above process is extended to a multi-head attention mechanism \( \text{MHA}(\cdot) \):
\begin{equation}
\text{MHA}(h_i) = \text{Concat}(\text{head}_1, \text{head}_2, \ldots, \text{head}_m) W_{\text{out}}
\end{equation}
where \( m \) represents the number of attention heads, \( \text{Concat}(\cdot) \) denotes the concatenation operation of the head vectors, and \( W_{\text{out}} \) is the output projection matrix. Each head is computed by \( W_{\text{query}} \), \( W_{\text{key}} \), and \( W_{\text{value}} \) according to the Formula~\ref{eqn:hi}.

\textbf{GNN Layer: }To further account for the graph structure and fine-grained differences between nodes, we combine the GNN layer with the Transformer layer to more deeply capture the complex interactions between drugs. For node \( v_i \), the message passing process in the GNN layer is as follows:
\begin{equation}
h_M(v_i) = \text{Message}(h_k, \{v_k\}_{v_k \in N(v_i)})
\end{equation}
\begin{equation}
h_i = \text{Combine}(h_i, h_M(\{v_i\}))
\end{equation}
Where \( N(v_i) \) denotes the set of neighbors of node \( v_i \). \( h_i \) and \( h_k \) represent the feature representations of node \( v_i \) and its neighbor node \( v_k \), respectively. \( \text{Message}(\cdot) \) is the message passing and aggregation function defined by the GNN layer, while \( \text{Combine}(\cdot) \) represents the method of information aggregation in the GNN layer.

\textbf{Sample Update Layer: }After the Transformer and GNN layers, the attention samples in the drug-microbe heterogeneous graph need to be adjusted according to the updated graph structure. Therefore, we use a message-passing-based attention sample update strategy: by leveraging the message-passing mechanism of the GNN layer, drug-microbe samples are directly updated without the need for calculating complex similarity matrices. Specifically, during the message passing phase of the GNN layer, we aggregate the attention information of the neighbors of each drug or microbe center node. These attention samples from the neighboring nodes may be related to the center node. We represent the set of attention samples from the neighboring nodes as attention messages, which are defined as follows:
\begin{equation}
\text{Attn\_Msg}(v_i) = U \text{Smp}(v_i) \{V_j\}_{v_j \in N(v_i)}
\end{equation}
where \( U \) is a learned weight matrix, \( \text{Smp}(v_i) \) represents the sample features of node \( v_i \), and \( \{V_j\}_{v_j \in N(v_i)} \) denotes the feature information of all neighboring nodes \( v_j \). This process helps to strengthen the focus on drug-microbe relationships and enhances the prediction ability.

\subsection{Divergence-Convergence feature enhancement strategy}
%After obtaining the drug and microbe features learned by the heterogeneous graph Transformer and similarity views, 
We design a feature enhancement strategy based on Divergence-Convergence. The "Divergence" phase enlarges the representation distance between features from different views through Adversarial Learning, preserving their differences. The "Convergence" phase uses Bidirectional Synergistic Attention Mechanism to aggregate multi-view features, achieving effective fusion in the feature space. Ablation experiments demonstrate the superiority of this approach. Just as repeatedly stretching dough makes noodles chewier, features that first diverge and then converge tend to have stronger expressive power. %Next, we will introduce the process through two phases: Divergence and Convergence.
\subsubsection{Divergence Phase}
The information bottleneck theory~\citep{hu2024survey} suggests that deep neural networks, during the learning process, should compress information through a bottleneck, eliminating noisy inputs and retaining only the most relevant features for general concepts. This enhances the network’s ability to extract crucial information. Inspired by this theory, in the Divergence phase, 
we designed Adversarial Learning method with the objective of increasing the feature space distance between the drug features from the drug-microbe association view ($d_{1,i}$) and drug similarity view ($d_{2,i}$), as well as between the microbe features derived from the drug-microbe association view ($d_{1,j}$) and the microbe similarity view ($d_{2,j}$), thereby strengthening the complementarity and diversity between the similarity information of drugs and microbes and the heterogeneous information.
% in the Divergence phase, we propose an Adversarial Learning method to establish a competitive relationship between the drug features from the drug association view ($d_{1,i}$) and drug similarity view ($d_{2,i}$), as well as between the microbe features derived from the microbe association view ($d_{1,j}$) and the microbe similarity view ($d_{2,j}$). 
% This Adversarial Learning method encourages mutual repulsion between the two representation spaces through distance maximization, which enhances feature diversity for subsequent fusion. 
The distance constraint formulation for drugs and microbes follows:
\begin{equation}
D_i = \| d_{1,i} - d_{2,i} \|_2
\end{equation}
\begin{equation}
D_j = \| d_{1,j} - d_{2,j} \|_2
\end{equation}
Building upon these distance metrics, we formulate the adversarial objective to strategically maximize the separation margin $\gamma$ between competing representations through dual-branch optimization:
\begin{equation}
\mathcal{L}_{adv,drug} = \frac{1}{N} \sum_{i=1}^{N} \max(0, \gamma - D_i)
\end{equation}
\begin{equation}
\mathcal{L}_{adv,microbe} = \frac{1}{N} \sum_{j=1}^{N} \max(0, \gamma - D_j)
\end{equation}
where $N$ denotes the total number of instances, and $\gamma$ controls the adversarial separation intensity. This mechanism drives the association-view and similarity-view representations to develop complementary patterns through strategic competition, enhancing information complementarity and diversity between association and similarity views.

\subsubsection{Convergence Phase}
In the Convergence phase, we propose a Bidirectional Synergistic Attention Mechanism (BSAM) to integrate complementary information from heterogeneous feature spaces. This architecture establishes dual-directional feature negotiation between association-view ($Z_1$) and similarity-view ($Z_2$) representations through weight-aware feature diplomacy, enabling adaptive information exchange while preserving view-specific characteristics.

The fusion process for drug features (analogously applied to microbes) operates through three strategic phases:

\textbf{Phase 1: Feature Diplomacy}
Establish communication channels between association features ($Z_1$) and similarity features ($Z_2$):
\begin{equation}
h_1 = \tanh(W_{\phi} z_1 + b_{\phi})
\end{equation}
\begin{equation}
h_2 = \tanh(W_{\psi} z_2 + b_{\psi})
\end{equation}
where $h_1, h_2$ denote the projected feature vectors, $W_{\phi}, W_{\psi}$ are learnable projection matrices, $b_{\phi}, b_{\psi}$ represent bias terms, and $\tanh(\cdot)$ indicates the hyperbolic tangent activation function.

\textbf{Phase 2: Synergy Discovery}
Compute view-specific compatibility scores through dual-perspective agreement:
\begin{equation}
e_1 = \mathbf{v}^T \cdot \text{ReLU}\left(W_{\omega}^{1}[h_1 \| h_2]\right)
\end{equation}
\begin{equation}
e_2 = \mathbf{v}^T \cdot \text{ReLU}\left(W_{\omega}^{2}[h_1 \| h_2]\right)
\end{equation}
where $W_{\omega}^{1}$ and $W_{\omega}^{2}$ are view-specific transformation matrices for association view and similarity view respectively, $\mathbf{v}$ is the shared attention vector, and $[h_1 \| h_2]$ denotes concatenated projected features.

\textbf{Phase 3: Consensus Formation}
Generate integrated features through normalized attention aggregation:
\begin{equation}
f = \sum_{i=1}^2 \frac{\exp(e_i)}{\sum_{j=1}^2 \exp(e_j)} \cdot z_i
\end{equation}

This three-phase synergistic fusion creates negotiated feature representations that preserve view-specific patterns while maximizing complementary information utilization, effectively addressing the feature redundancy problem in conventional multi-view fusion approaches.

\subsection{Drug-microbe association prediction}
We input the fused features into a three-layer multilayer perceptron (MLP) to obtain the score label, indicating whether the drug and microbe match (positive or negative sample):
\begin{equation}
\hat{y}_{ij} = \text{MLP}(f_r^i \parallel f_d^j)
\end{equation}
where \( \hat{y}_{ij} \) represents the likelihood of the association between drug \( i \) and microbe \( j \). \( f_r^i \) and \( f_d^j \) denote the fused features of the drug and microbe, respectively.

In the overall loss function, we use binary cross-entropy loss as the primary loss, and the formula is as follows:
\begin{equation}
L_{rel} = -\sum_{(i,j)} \left[ \mathbf{W}^* \mathbf{y}_{ij} \log(\sigma(z_i)) + (1 - \mathbf{y}_{ij}) \log(\sigma(z_i)) \right]
\end{equation}
where \( (i, j) \) represents the pair of drug \( i \) and microbe \( j \), \( w \) is the weight for positive samples, \( \sigma(z_i) \) is the Sigmoid activation function, and \( Z_i \) is the raw predicted score before the activation function.

The overall loss function is the weighted sum of the binary cross-entropy loss and the contrastive loss:
\begin{equation}
L_{\text{total}} = L_{\text{rel}} + \beta_1 L_{\text{con,drug}} + \beta_2 L_{\text{con,mico}}
\end{equation}
where \( \beta_1\) and \( \beta_2 \) are hyperparameters that control the weight of the two contrastive losses in the total loss, balancing the proportion of the scoring prediction task and the adversarial learning task, as well as the weight of the drug and microbe scoring learning tasks.

\section{Experiment}
In this study, we utilized association data from the Microbe-Drug Association Database (MDAD)~\citep{sun2018mdad}. Drug-drug similarity, microbe-microbe similarity, and drug-microbe associations, are derived from ~\citep{xuan2024multi}, including 173 microbes, 1373 drugs and their associations. The final drug similarity was obtained by the weighted sum of Gaussian kernel similarity and drug structural similarity. The complete dataset was randomly split into a training set and a testing set with the testing set representing 10\% of the total data. 

\subsection{Baseline model}
To assess the performance of the proposed model, we compare DCFA\_DMP with four publicly available drug-microbe prediction methods listed below:
 SCSMDA~\citep{tian2023predicting}, NGMDA~\citep{xuan2024multi}, DHDMP~\citep{xuan2024dynamic}, and GACNNMDA~\citep{ma2023gacnnmda}. Each of these methods was trained using the optimal parameters mentioned in their respective original papers. To ensure consistency, we trained DCFA\_DMP and the other methods on the same dataset. The descriptions of these four comparison methods are provided in Section~\ref{tab:Introduction}.

\begin{table*}[htbp]
\centering
\caption{Performance metrics of different baseline methods on independent test set.}
\begin{tabular}{>{\centering\arraybackslash}p{1.6cm} >{\centering\arraybackslash}p{1.6cm} >{\centering\arraybackslash}p{1.6cm} >{\centering\arraybackslash}p{1.6cm} >{\centering\arraybackslash}p{1.6cm} >{\centering\arraybackslash}p{1.6cm} >
{\centering\arraybackslash}p{2.5cm} >
{\centering\arraybackslash}p{2.3cm}
}
\toprule
Methods & AUROC & AUPR & Recall & F1-score & Precision & P-value of AUROCs & P-value of AUPRs\\
\midrule
SCSMDA & 0.7642±0.0455 & \underline{0.7276±0.0551} & 0.7695±0.0659 & \underline{0.7281±0.0373} & \underline{0.6865±0.0216} &
1.26E-02 &
2.98E-11
\\
NGMDA & \underline{0.9761±0.0029} & 0.5646±0.0472 & \textbf{0.9369±0.0074} & 0.0556±0.0138 & 0.0285±0.0069 & 3.84E-02 & 1.99E-03\\
DHDMP & 0.9736±0.0044 & 0.7033±0.0906 & \underline{0.8557±0.0363} & 0.245±0.0713 & 0.1449±0.0500 & 5.97E-08 & 1.71E-08\\
GACNNMDA & 0.9742±0.0073 & 0.6937±0.1153 & 0.4130±0.0467 & 0.3916±0.1910 & 0.3723±0.2025 & 3.13E-02 & 5.18E-04\\
DCFA\_DMP & \textbf{0.9894±0.0063} & \textbf{0.9856±0.0162} & 0.6696±0.0371 & \textbf{0.7992±0.0275} & \textbf{0.9920±0.0125} & \texttt{--} & \texttt{--}\\
\bottomrule
\end{tabular}
\captionsetup{font=scriptsize}
\label{tab:performance_metrics}
\end{table*}
\begin{figure*}
\centering
\includegraphics[width=\textwidth]{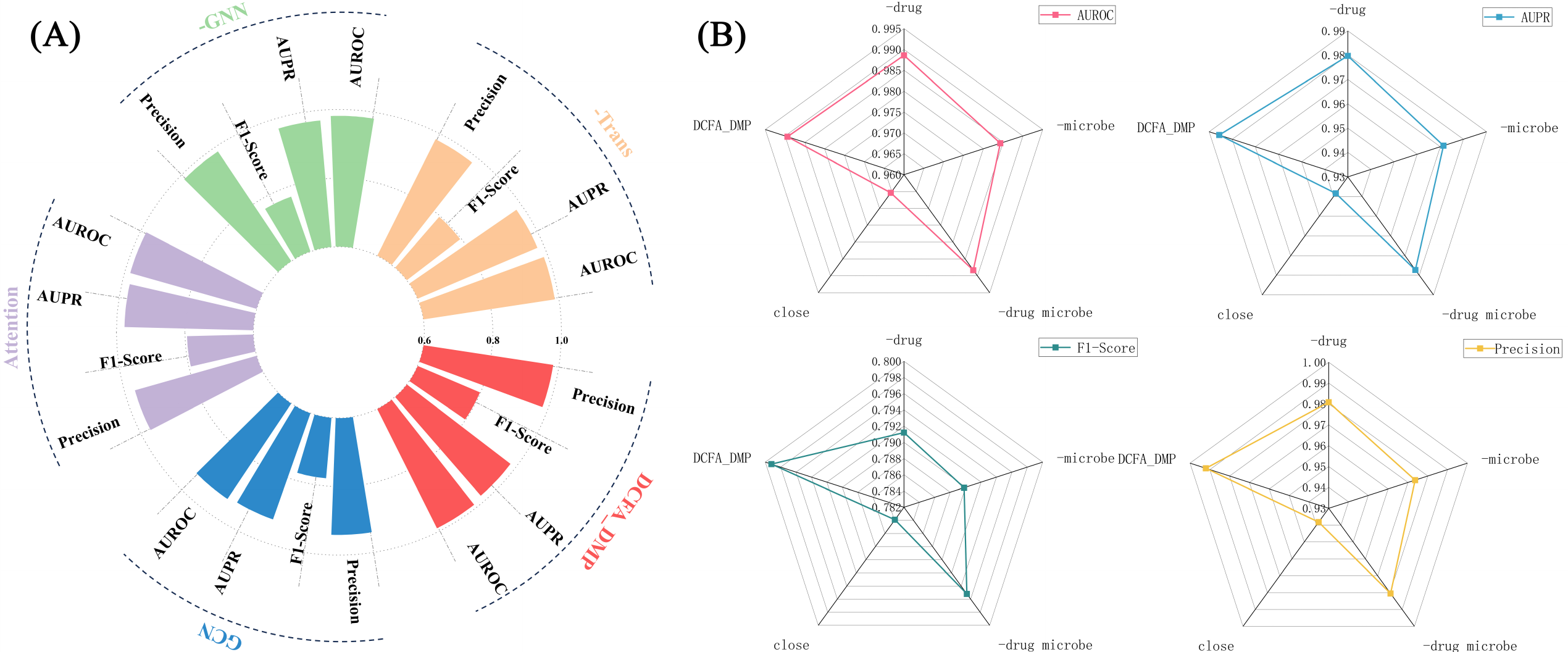}
\captionsetup{font=scriptsize}
\caption{Ablation results for (A) Transformer-based Graph; (B) Adversarial Learning method.}
\label{fig:4}
\end{figure*}

As shown in Table~\ref{tab:performance_metrics}, DCFA\_DMP not only significantly outperforms the current state-of-the-art (SOTA) methods in metrics such as AUROC and AUPR but also demonstrates optimal performance in terms of positive-negative sample balance, as evidenced by the comprehensive evaluation metric F1.

To further investigate whether the predictive performance of DCFA\_DMP is significantly higher than that of each comparison method, we performed statistical tests on the AUROC and AUPR values of these methods. The results are shown in Table~\ref{tab:performance_metrics}. It can be seen that the p-values for all four methods in terms of AUROC and AUPR are less than 0.05, indicating statistically significant differences, which further corroborates the superior predictive performance of DCFA\_DMP.

\subsection{Ablation study}
Given that DCFA\_DMP mainly consists of three components: Transformer-based Graph,  Adversarial Learning method and BSAM, we conducted ablation experiments on these components.
\subsubsection{Ablation of Transformer-based Graph}
We set up four scenarios to investigate the impact of the transformer-based graph on the model, namely: without the transformer component (-Trans), without the GNN component (-GNN), replacing the transformer component with a standard attention mechanism (Attention), and replacing the GNN component with GCN (GCN). The results are shown in Figures~\ref{fig:4} (A). When the Transformer layer (-Trans) or GNN layer (-GNN) is removed, performance decreases, which demonstrates the necessity of both components. Their capabilities complement each other, enhancing the learning of feature information. Replacing the Transformer with a standard attention mechanism or the GNN with a GCN also results in a significant performance drop, highlighting the superiority of the Transformer and GNN in handling heterogeneous network data. Transformer uses the self-attention mechanism to effectively capture global information, and by introducing multi-head attention, it learns information from different subspaces, enhancing the expressiveness of the model and enabling deeper feature representations. In contrast, the traditional GCN, primarily designed for homogeneous graphs, has limited capacity to process multiple node and edge types, while the GNN captures finer graph structural details.
\begin{figure*}
\centering
\includegraphics[width=\textwidth]{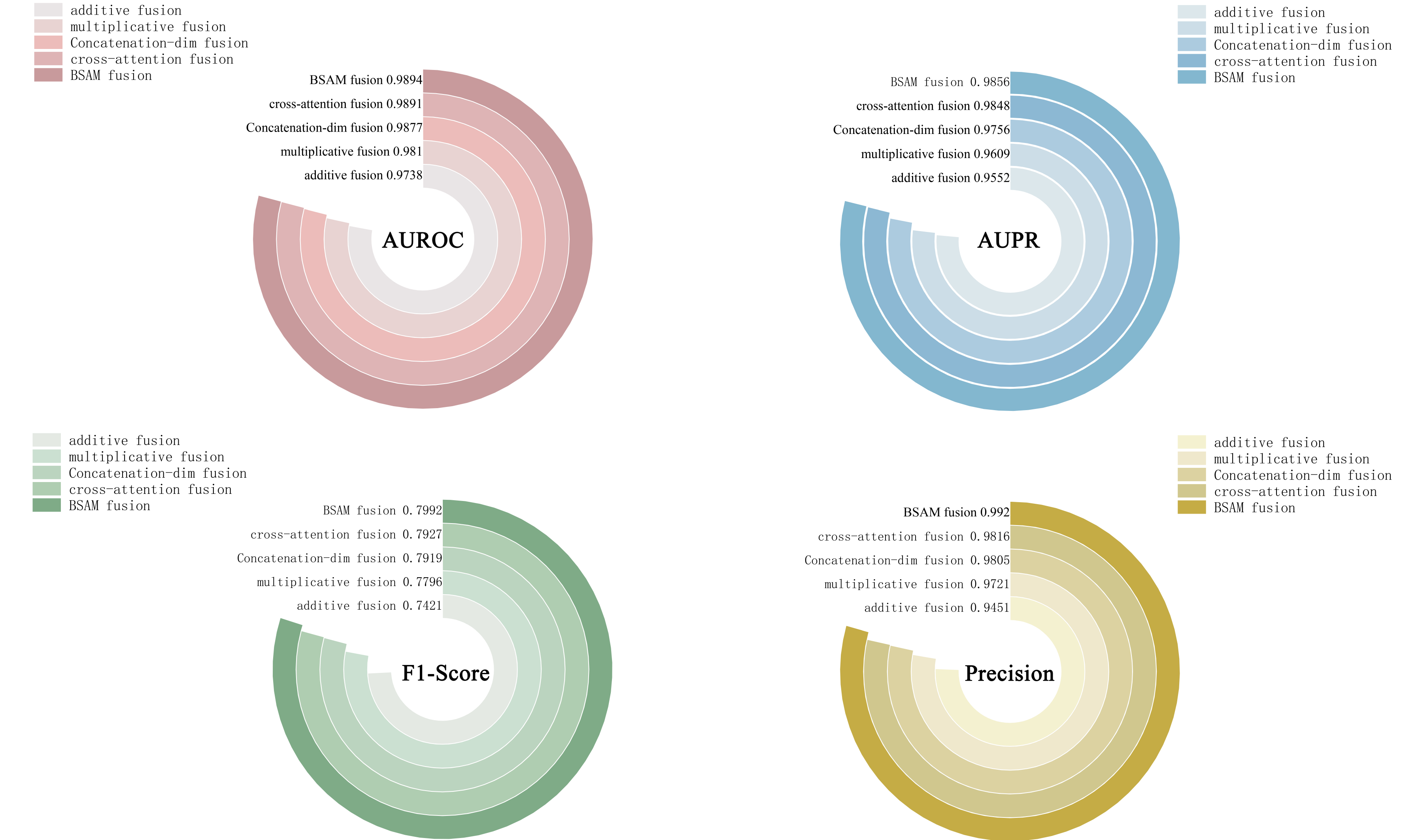}
\captionsetup{font=scriptsize}
\caption{Ablation results for Bidirectional Synergistic Attention Mechanism.}
\label{fig:4.5}
\end{figure*}
\subsubsection{Ablation of Adversarial Learning method} 
We set up four scenarios to investigate the impact of the Adversarial Learning method on the model, namely: without drug adversarial learning (-drug), without microbe adversarial learning (-micro), without drug and microbe adversarial learning (-drug microbe), and replacing the feature divergence in adversarial learning with feature convergence (close), as shown in Figures~\ref{fig:4} (B).
% The main conclusions we draw are as follows:
The divergence of drug and microbe features through adversarial learning significantly impacts feature enhancement. When the features are brought closer together, the feature differences between drug or microbe nodes become smaller, and the useful information learned by Bidirectional Synergistic Attention Mechanism decreases, thereby reducing performance. The performance decline when drug adversarial learning, microbe adversarial learning, or both are removed further highlights the importance of adversarial learning in the feature fusion process.
\subsubsection{Ablation of Bidirectional Synergistic Attention Mechanism} 
To systematically evaluate the efficacy of the Bidirectional Synergistic Attention Mechanism (BSAM) in the convergence phase, we compared it with four baseline fusion strategies: additive fusion, multiplicative fusion, concatenation followed by dimensionality reduction-based fusion (concatenation-dim fusion) , and cross-attention fusion. As illustrated in Figures~\ref{fig:4.5}, BSAM achieved the highest performance across all four evaluation metrics. The proposed BSAM effectively facilitates the synergistic integration of complementary features across different views, enabling comprehensive multi-dimensional interactions and enhanced fusion of multi-view characteristics. The experimental results conclusively validate the superior fusion efficacy of BSAM, offering a robust solution for complex multi-view learning tasks requiring discriminative feature integration.

\begin{figure*}
\centering
\includegraphics[width=\textwidth]{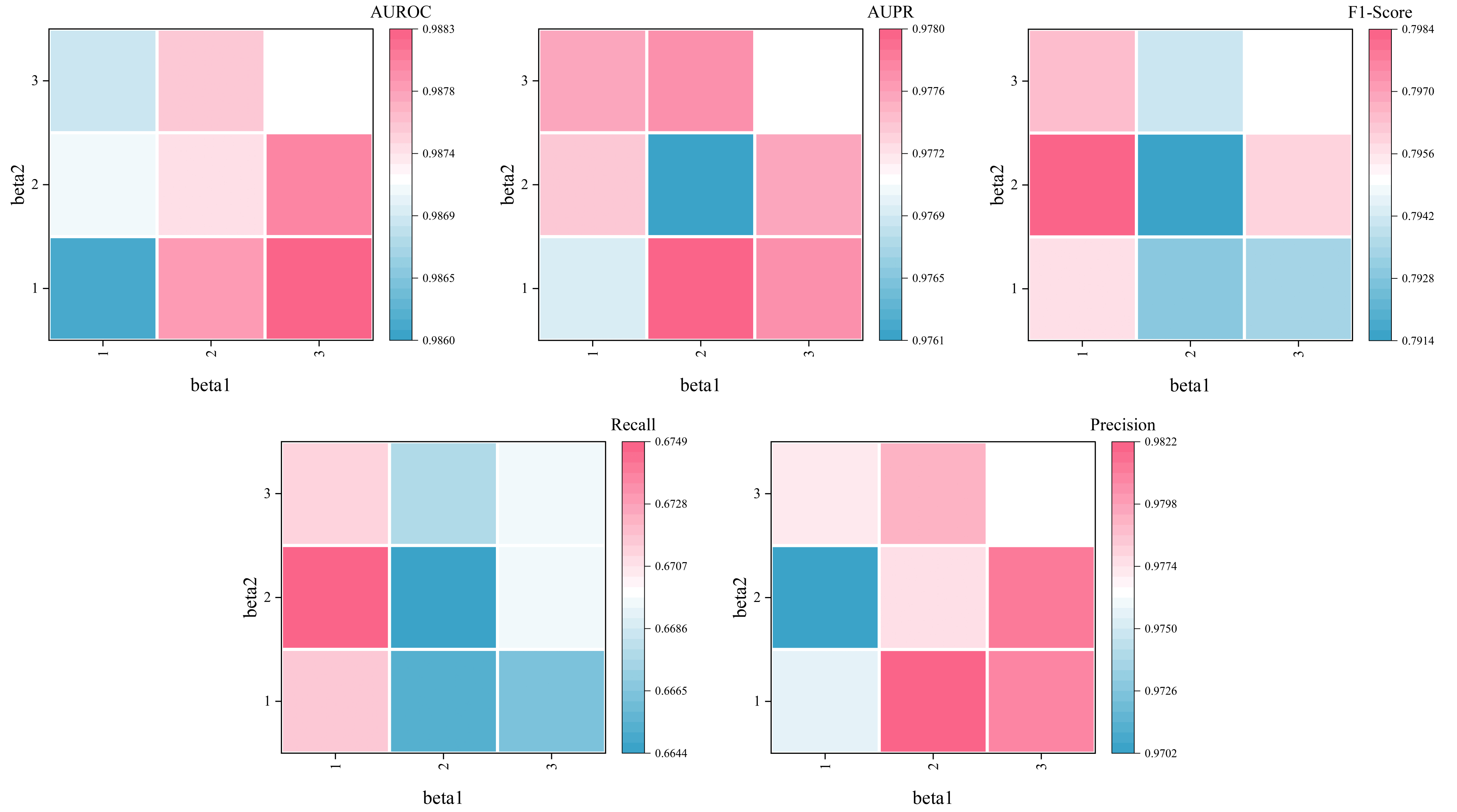}
\captionsetup{font=scriptsize}
\caption{Parameter analysis for different trade-off parameters (\( \beta_1 \) and \( \beta_2 \)) values.}
\label{fig:2}
\end{figure*}
\begin{figure*}
\centering
\includegraphics[width=\textwidth]{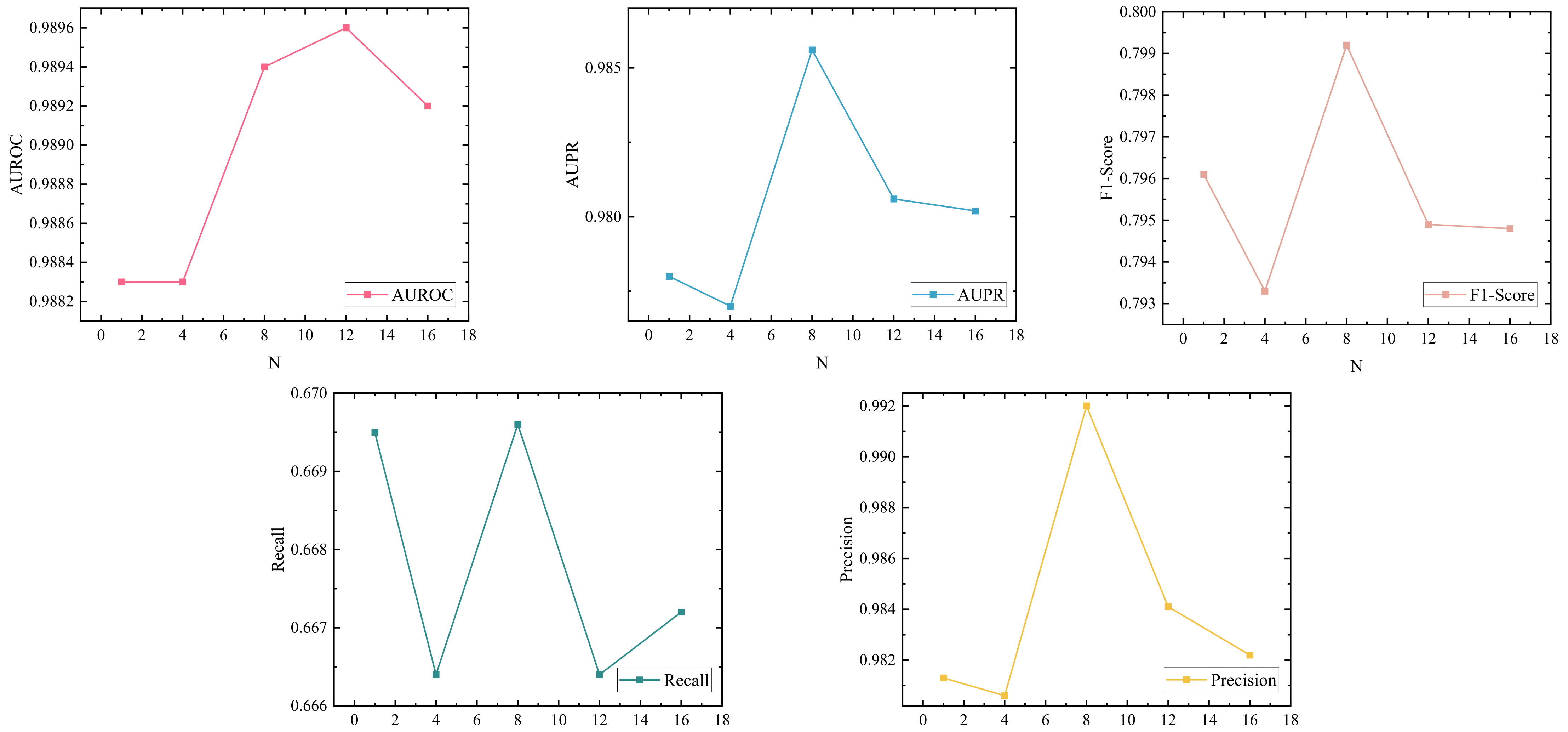}
\captionsetup{font=scriptsize}
\caption{Parameter analysis for different number of neighboring nodes.}
\label{fig:3}
\end{figure*}
\subsection{Parameter sensitivity analysis}
In our model, several important parameters are considered, including the learning rate \( \text{lr} \), the number of training epochs \( a \), dropout rate, the trade-off parameters \( \beta_1 \) and \( \beta_2 \), and the number of neighbors \( N \) in the feature graph. 
% We explored different combinations of these parameters, with some of the value ranges as follows: \( \text{lr} \in \{0.001, 0.005, 0.01, 0.1\} \); \( a \in \{1000, 2000, 3000, 4000\} \); dropout \( \in \{0.1, 0.2, 0.3, 0.4, 0.5\} \). 
Through empirical tuning of these parameters, we set \( a = 4000 \), \( \text{lr} = 0.005 \), and dropout = 0.5 for all experiments.
\subsubsection{Trade-off parameters analysis}
We further conducted experimental verification on the impact of the trade-off parameters \( \beta_1 \) and \( \beta_2 \) as well as the number of neighbors \( N \) in the feature graph across all datasets. The trade-off parameters are used to appropriately balance the relative importance of adversarial learning and scoring prediction tasks. By setting both trade-off parameters, we can also balance the contributions of the drug and microbe adversarial learning tasks. The values of the trade-off parameters \( \beta_1 \) and \( \beta_2 \) were chosen from \{0.01, 0.02, 0.03\}, and all possible pairwise combinations were tested. Figure~\ref{fig:2} shows the results of various metrics under different trade-off parameters. Considering the performance across five metrics, the optimal overall results were achieved when both \( \beta_1 \) and \( \beta_2 \) were set to 0.03. Therefore, in our method, we set both \( \beta_1 \) and \( \beta_2 \) to 0.03 as the model parameters.
\begin{figure*}
\centering
\includegraphics[width=\textwidth]{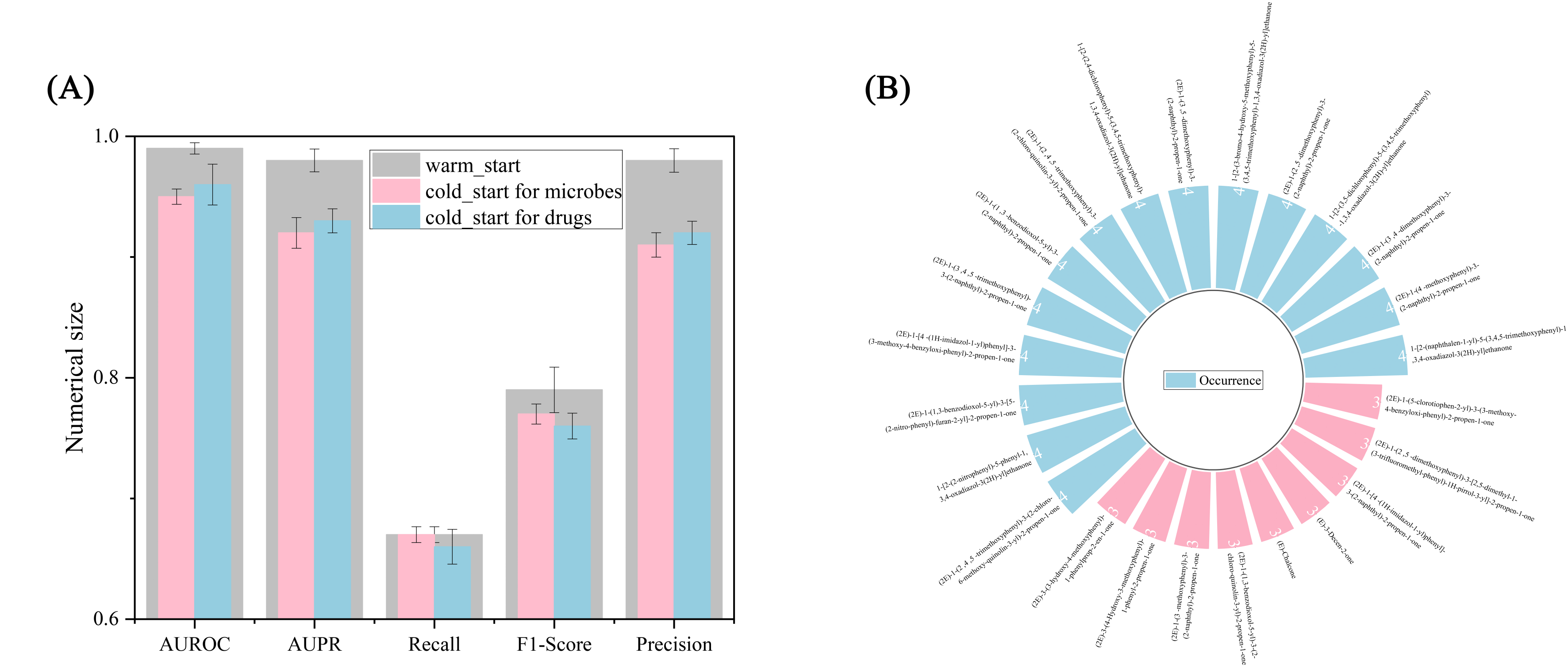}
\captionsetup{font=scriptsize}
\caption{(A) Results of performance metrics under cold start condition. (B) The top 25\% of potential drugs predicted by DCFA\_DMP for the treatment of COVID-19.}
\label{fig:6}
\end{figure*}
\subsubsection{Neighboring nodes analysis}
The number of neighboring nodes \( N \) in the feature graph influences the node update mechanism and its scope, affecting the model's performance. We analyzed the stability of our method on the dataset by varying \( N \). The results regarding the impact of the number of neighbors are shown in Figure~\ref{fig:3}. Intuitively, we set the range of \( N \) to \{1, 4, 8, 12, 16\}. As shown in the figure, when the number of neighbors in the feature space is set to \( N = 8 \), the model achieves the optimal overall performance. When \( N \) is smaller than 8, the model fails to capture sufficient global or deeper information. Conversely, when \( N \) exceeds 8, it leads to an over-smoothing issue, causing the representations of different types of nodes to converge, thereby reducing the model's discriminative ability.
\subsection{Cold-start study}
Despite significant progress in predicting the drug-microbe relationships in recent years, many methods heavily rely on existing data and suffer from the "cold start" problem when identifying new drugs or microbes. Therefore, we conducted a "cold start" experiment to evaluate the ability of DCFA\_DMP to identify new drugs and microbes. Specifically, for each drug \( r_i \), we removed all known drug-disease associations involving \( r_i \) as the test set, using the remaining associations for training. A similar approach was applied to microbes. Given that the dataset contains 1373 drugs and 173 microbes, we selected 2\% and 4\% of these for testing, and averaged the results. The outcomes are shown in Figure~\ref{fig:6} (A). 

Compared to the warm start, although the cold start performance for both drugs and microbes declined, the prediction results remained promising. For the prediction of new microbes, the AUROC was approximately 0.95, AUPR was about 0.92, Recall was around 0.67, F1-Score was approximately 0.77, and Precision was about 0.91. For the prediction of new drugs, AUROC was approximately 0.96, AUPR was around 0.93, Recall was about 0.66, F1-Score was approximately 0.76, and Precision was about 0.92. The prediction results for new drugs were mostly higher than those for new microbes, likely because the dataset contains a much larger number of drugs than microbes, resulting in richer drug features and better prediction performance. This demonstrates that DCFA\_DMP still exhibits significant effectiveness even when the association information between drugs and microbes is extremely limited.

% \begin{figure}[htbp]
% \centering
% \begin{minipage}{0.48\textwidth}
%   \centering
%   \includegraphics[width=2.5in]{Graph15.pdf}
%   \captionsetup{font=scriptsize}
%   \caption{Results of performance metrics under cold start condition.}
%   \label{fig:6}
% \end{minipage}%
% \hfill
% \begin{minipage}{0.48\textwidth}
%   \centering
%   \includegraphics[width=2.8in]{G13.pdf}
%   \captionsetup{font=scriptsize}
%   \caption{Predict the top 25\% of potential associated drugs.}
%   \label{fig:7}
% \end{minipage}
% \end{figure}

\subsection{Case study}
To further validate the reliability of DCFA\_DMP in real-world drug-microbe association prediction, we conducted a case study. Given the recurring nature of the COVID-19 pandemic, predicting potential drugs for treatment is of great importance. From the microbial perspective, we aimed to identify possible drugs for COVID-19.

First, we collected twelve microbes associated with COVID-19 from published literature and used the relationship dataset recorded in the MDAD database as the training dataset for prediction. We ranked the drugs based on the association scores predicted by the model and selected the top 20 drugs. Next, to account for the possibility of predicting duplicate drugs, where the higher the frequency of prediction, the more likely the drug is to be effective in treating COVID-19, we ranked all the predicted drugs by their occurrence frequency and selected the top 25\% for analysis. The results are shown in the Figure~\ref{fig:6} (B). 

For (E)-3-Decen-2-one (E3D2o), the only drug not directly validated in the dataset, we found its potential for COVID-19 treatment in the study by Javier et al~\citep{asensio2024multimodal}. Their research indicates that biofilm formation by \textit{Haemophilus influenzae} is closely associated with human nasopharyngeal colonization, pediatric otitis media, and chronic respiratory infections in adults, while E3D2o demonstrates effectiveness in inhibiting biofilm formation and eradicating pre-formed biofilms of \textit{H. influenzae}. Furthermore, the dataset reveals that E3D2o is known to be associated with two microorganisms, \textit{Vibrio cholerae}~\citep{crisan2020vibrio} (an intestinal pathogen that induces intestinal inflammation) and \textit{Vibrio harveyi}~\citep{yu2022vibrio} (which triggers inflammation by inducing the secretion of pro-inflammatory cytokines). This further supports the potential application of E3D2o in the treatment of COVID-19.

\section{Conclusion}
In this paper, we propose a multi-view divergence-convergence feature augmentation framework for drug-related microbes prediction, named DCFA\_DMP, for predicting potential drug-microbe associations. This model enhances the information complementarity and diversity between the association and similarity views through Adversarial Learning, and introduces a novel BSAM to aggregate multi-view features, thereby facilitating interactive communication and fusion of heterogeneous information. For the heterogeneous network, we first use a Transformer to expand the receptive field of the GNN. Then, through the GNN, we learn the most relevant information of the nodes, and during the message-passing process of the GNN, we perform drug-microbe sample updates to reduce computational complexity. Experiments demonstrate that DCFA\_DMP exhibits significant effectiveness and shows robustness and reliability in cold-start scenarios as well as in real case studies.

In the future, we can incorporate richer biological information to enrich the feature networks, such as drug-drug interactions, microbe-microbe interactions, and disease-microbe associations, and explore applying the model to other association prediction tasks, such as RNA-disease association prediction.
\section{Acknowledgement}
This work has been supported by the National Natural Science Foundation of China (62302075, 62472062), the Innovation Support Program for Dalian High-Level Talents (2023RQ007), the Dalian Excellent Young Project (2022RY35), and the Dalian Science and technology Innovation Fund project (2024JJ12GX022).

% In the future, we plan to improve MHMDA's association prediction in several ways: (1) incorporating more diverse biological association information, such as lncRNA-disease, lncRNA-miRNA, circRNA-disease, and circRNA-miRNA associations, into the model; (2) reducing the model's reliance on similarity data and minimizing the impact of similarity computation on the model, possibly through methods like Semi-supervised Generative Adversarial Networks (SGAN); 

% \section{Key Points}\label{sec5}
% \boxedtext{
% \begin{itemize}
% \item We first propose a biologically interpretable "similarity-assiciation-similarity" structure multi-hop Metapath learning method based on hierarchical attention, aiming to unearth potential association information on specific long paths between miRNAs and diseases.
% % \item We proposed a novel graph neural network heterohypernet (HeteroHyperNet) that integrates heterogeneous features and hyper-features, focusing on comprehensively exploring both known and potential associations between miRNAs and diseases.
% \item We propose a novel graph neural network HeteroHyperNet that integrates heterogeneous real edges and virtual edges, focusing on comprehensively exploring both known and potential associations between miRNAs and diseases.
% \item We conduct extensive experiments in the miRNA-disease association testing scenario, involving both warm-start and cold-start approaches. The results demonstrate that MHMDA attains commendable success in miRNA-disease association prediction. 
% \end{itemize}}

\bibliographystyle{unsrtnat}
\bibliography{tcbb}

\end{document}